\documentclass[10pt, a4paper]{article}
\usepackage{lrec}
\usepackage{multibib}
\newcites{languageresource}{Language Resources}
\usepackage{lipsum}
\usepackage{graphicx}
\usepackage{tabularx}
\usepackage{soul}
\usepackage{subfig}
\usepackage{epstopdf}
\usepackage[utf8]{inputenc}
\usepackage{hyperref}
\usepackage{xstring}

\usepackage{color}

\title{Corpus Creation for Sentiment Analysis in Code-Mixed Tamil-English Text}

\name{\begin{tabular}{c}Bharathi Raja Chakravarthi\(^1\),Vigneshwaran Muralidaran\(^2\), \\ Ruba Priyadharshini\(^3\), John P. McCrae\(^1\) \end{tabular}}

\address{ \(^1\)Insight SFI Research Centre for Data Analytics, Data Science Institute, \\ National University of Ireland Galway,  \{bharathi.raja, john.mccrae\}@insight-centre.org\\
 \(^2\)School of English, Communication and Philosophy, Cardiff University, muralidaranV@cardiff.ac.uk\\
\(^3\)Saraswathi Narayanan College, Madurai, India, rubapriyadharshini.a@gmail.com\\}

\abstract{
Understanding the sentiment of a comment from a video or an image is an essential task in many applications. Sentiment analysis of a text can be useful for various decision-making processes. One such application is to analyse the popular sentiments of videos on social media based on viewer comments. However, comments from social media do not follow strict rules of grammar, and they contain mixing of more than one language, often written in non-native scripts. Non-availability of annotated code-mixed data for a low-resourced language like Tamil also adds difficulty to this problem. To overcome this, we created a gold standard Tamil-English code-switched, sentiment-annotated corpus containing 15,744 comment posts from YouTube. In this paper, we describe the process of creating the corpus and assigning polarities. We present inter-annotator agreement and show the results of sentiment analysis trained on this corpus as a benchmark. \\ \newline  }
\begin{document}
\maketitleabstract
\section{Introduction}
Sentiment analysis has become important in social media research  \cite{yang-eisenstein-2017-overcoming}. Until recently these applications were created for high-resourced languages which analysed monolingual utterances. But social media in multilingual communities contains more code-mixed text \cite{barman-etal-2014-code,chanda-etal-2016-unraveling,pratapa-etal-2018-language,winata-etal-2019-learning}. Our study focuses on sentiment analysis in Tamil, which has little annotated data for code-mixed scenarios \cite{phani-etal-2016-sentiment,chakravarthi-code-mix-survey}. Features based on the lexical properties such as a dictionary of words and parts of speech tagging have less performance compared to the supervised learning \cite{kannan-etal-2016-towards} approaches using annotated data. However, an annotated corpus developed for monolingual data cannot deal with code-mixed usage and therefore it fails to yield good results \cite{alghamdi-etal-2016-part,ws-2018-approaches} due to mixture of languages at different levels of linguistic analysis. \\

Code-mixing is common among speakers in a bilingual speech community. As English is seen as the language of prestige and education, the influence of lexicon, connectives and phrases from English language is common in spoken Tamil. It is largely observed in educated speakers although not completely absent amongst less educated and uneducated speakers \cite{krishnasamy2015code}. Due to their pervasiveness of English online, code-mixed Tamil-English (Tanglish) sentences are often typed in Roman script \cite{suryawanshi-etal-2020-meme,suryawanshi-etal-2020-tamil-meme}. \\

We present TamilMixSentiment \footnote{https://github.com/bharathichezhiyan/TamilMixSentiment}, a dataset of YouTube video comments in Tanglish. TamilMixSentiment was developed with guidelines following the work of \newcite{mohammad-2016-practical} and without annotating the word level language tag. The instructions enabled light and speedy annotation while maintaining consistency. The overall inter-annotator agreement in terms of Kripendorffs's $\alpha$ \cite{krippendorff5} stands at 0.6. In total, 15,744 comments were annotated; this makes the largest general domain sentiment dataset for this relatively low-resource language with code-mixing phenomenon. \\

We observed all the three types of code-mixed sentences \-- Inter-Sentential switch, Intra-Sentential switch and Tag switching. Most comments were written in Roman script with either Tamil grammar with English lexicon or English grammar with Tamil lexicon. Some comments were written in Tamil script with English expressions in between. The following examples illustrate the point.

\begin{itemize}
    \item {\color{blue}\textbf{Intha padam vantha piragu yellarum Thala ya kondaduvanga.} }\-- \textit{After the movie release, everybody will celebrate the hero.} Tamil words written in Roman script with no English switch.
    \item \textbf{{\color{blue}{\color{red}Trailer late} ah parthavanga {\color{red}like} podunga.}} \-- \textit{Those who watched the trailer late, please like it.} Tag switching with English words.
    \item \textbf{{\color{red}Omg .. use head phones. }{\color{blue}Enna {\color{red}bgm} da saami ..}} \-- \textit{OMG! Use your headphones. Good Lord, What a background score!} Inter-sentential switch
    \item \textbf{{\color{red}I think {\color{blue}sivakarthickku} hero getup set {\color{blue}aagala.}}} \-- \textit{I think the hero role does not suit Sivakarthick.} Intra-sentential switch between clauses.
\end{itemize}

In this work we present our dataset, annotation scheme and investigate the properties and statistics of the dataset and information about the annotators. We also present baseline classification results on the new dataset with ten models to establish a baseline for future comparisons. The best results were achieved with models that use logistic regression and random forest.\\

The contribution of this paper is two-fold:
\begin{enumerate}
  \item We present the first gold standard code-mixed Tamil-English dataset annotated for sentiment analysis.
\item We provide an experimental analysis of logistic regression, naive Bayes, decision tree, random forest, SVM, dynamic meta-embedding, contextualized dynamic meta-embedding, 1DConv-LSTM and BERT on our code-mixed data for sentiment classification.
\end{enumerate}
\section{Related Work}
Recently, there has been a considerable amount of work and effort to collect resources for code-switched text. However, code-switched datasets and lexicons for sentiment analysis are still limited in number, size and availability. For monolingual analysis, there exist various corpora for English \cite{10.1145/1014052.1014073,Wiebe2005,jiang-etal-2019-challenge}, Russian \cite{rogers-etal-2018-rusentiment}, German \cite{cieliebak-etal-2017-twitter}, Norwegian \cite{maehlum-etal-2019-annotating} and Indian languages \cite{agrawal-etal-2018-beating,priya-etal-2020-senti-comparative}. 

When it comes to code-mixing, an English-Hindi corpus was created by \cite{sitaramsenti,joshi-etal-2016-towards,patra2018sentiment}, an English-Spanish corpus was introduced by \cite{solorio-etal-2014-overview,vilares-etal-2015-sentiment,VILARES16.43}, and a Chinese-English one \cite{lee-wang-2015-emotion} was collected from Weibo.com and English-Bengali data were released by Patra et al. \cite{patra2018sentiment}. 

Tamil is a Dravidian language spoken by Tamil people in India, Sri Lanka and by the Tamil diaspora around the world, with official recognition in India, Sri Lanka and Singapore \cite{chakravarthi2018improving,chakravarthi2019comparison,chakravarthi-etal-2019-wordnet,chakravarthi-etal-2019-multilingual}. Several research activities on sentiment analysis in Tamil \cite{8089122} and other Indian languages \cite{prakas-bharathi,das-bandyopadhyay-2010-sentiwordnet,a-r-etal-2012-cross,phani-etal-2016-sentiment,7880246,chakravarthi-code-mix-ruba-ne,chakravarthi-etal-2020-senti-malayalam} are happening because the sheer number of native speakers are a potential market for commercial NLP applications. However, sentiment analysis on  Tamil-English code-mixed data \cite{patra2018sentiment} is under-developed and data tare not readily available for research.

Until recently, word-level annotations were used for research in code-mixed corpora. Almost all the previous systems proposed were based on data annotated at the word-level. This is not only time-consuming but also expensive to create. However, neural networks and meta-embeddings \cite{kiela-etal-2018-dynamic} have shown great promise in code-switched research without the need for word-level annotation. In particular, work by \newcite{winata-etal-2019-learning} learns to utilise information from pre-trained embeddings without explicit word-level language tags. A recent work by \newcite{winata-etal-2019-hierarchical} utilised the subword-level information from closely related languages to improve the performance on the code-mixed text. \\

As there was no previous dataset available for Tamil-English (Tanglish) sentiment annotation, we create a sentiment dataset for Tanglish with voluntary annotators. We also show the baseline results with a few models explained in Section \ref{benchmark_results} 

\begin{figure}[ht] 
  \centering
  \subfloat[Example 1]{\includegraphics[width=0.4\textwidth,     height=12cm]{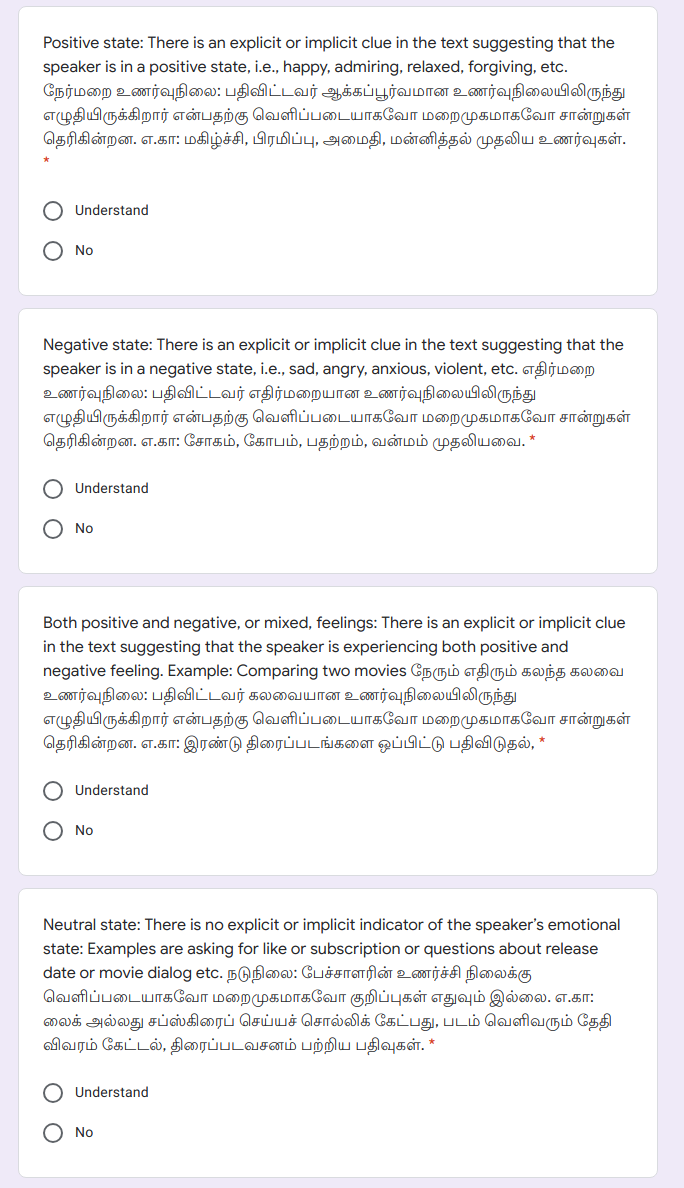}\label{fig:Images/f1}}
  \hfill
  \subfloat[Example 2]{\includegraphics[width=0.4\textwidth,     height=7cm]{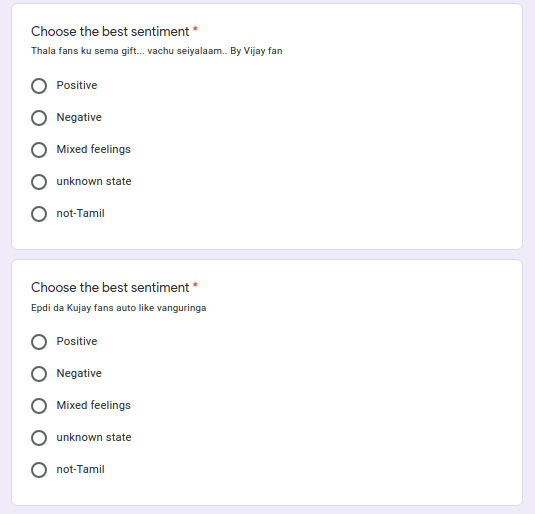}\label{fig:f2}}
  \caption{Examples of Google Form.}
  \label{fig:annoat}
\end{figure}

\section{Corpus Creation and Annotation}
Our goal was to create a code-mixed dataset for Tamil to ensure that enough data are available for research purposes.  We used the \textit{YouTube Comment Scraper tool}\footnote{https://github.com/philbot9/youtube-comment-scraper} and collected 184,573 sentences for Tamil from YouTube comments. We collected the comments from the trailers of a movies released in 2019. Many of the them contained sentences that were either entirely written in English or code-mixed Tamil-English or fully written in Tamil. So we filtered out a non-code-mixed corpus based on language identification at comment level using the \textit{langdetect library} \footnote{https://pypi.org/project/langdetect/}. Thus if the comment is written fully in Tamil or English, we discarded that comment since monolingual resources are available for these languages. We also identified if the sentences were written in other languages such as Hindi, Malayalam, Urdu, Telugu, and Kannada. We preprocessed the comments by removing the emoticons and applying a sentence length filter. 
We want to create a code-mixed corpus of reasonable size with sentences that have fairly defined sentiments which will be useful for future research. Thus our filter removed sentences with less than five words and more than 15 words after cleaning the data. In the end we got 15,744 Tanglish sentences.

\subsection{Annotation Setup}
For annotation, we adopted the approach taken by \newcite{mohammad-2016-practical}, and a minimum of three annotators annotated each sentence in the dataset according to the following schema shown in the Figure \ref{fig:annoat}. We added new category \textbf{Other language:} If the sentence is written in some other language other than Tamil or English. Examples for this are the comments written in other Indian languages using the Roman script.  
The annotation guidelines are given in English and Tamil. \\

As we have collected data from YouTube we anonymized to keep the privacy of the users who commented on it. As the voluntary annotators' personal information were collected to know about the them, this gives rise to both ethical, privacy and legal concerns. Therefore, the annotators were informed in the beginning that their data is being recorded and they can choose to withdraw from the process at any stage of annotation. The annotators should actively agree to being recorded. We created Google Forms in which we collected the annotators' email addresses which we used to ensure that an annotator was allowed to label a given sentence only once. We collected the information on gender, education and medium of instruction in school to know the diversity of annotators. Each Google form has been set to contain a maximum of 100 sentences. Example of the Google form is given in the Figure \ref{fig:annoat}. The annotators have to agree that they understood the scheme; otherwise, they cannot proceed further. Three steps complete the annotation setup. First, each sentence was annotated by two people. In the second step, the data were collected if both of them agreed. In the case of conflict, a third person annotated the sentence. In the third step, if all the three of them did not agree, then two more annotators annotated the sentences. 
\begin{table}[!htb]  
\begin{center} 
\renewcommand{\tabcolsep}{1.5mm}
\captionsetup{font=normalsize}
\normalsize
\begin{tabular}{|l|l|l|}
\hline
Gender & Male  & 9 \\
& Female  & 2 \\
\hline
Higher Education & Undegraduate  & 2 \\
& Graduate  & 2 \\
& Postgraduate  & 7 \\
\hline
Medium of Schooling & English  & 6 \\
& Tamil  & 5 \\
\hline
Total & &11\\
\hline
\end{tabular}
\caption{Annotators } 
\label{tab:annotators} 
\end{center} 
\end{table}
\subsection{Annotators}
To control the quality of annotation, we removed the annotator who did not annotate well in the first form. For example, if the annotators showed unreasonable delay in responding or if they labelled all sentences with the same sentiment or if more than fifty annotations in a form were wrong, we removed those contributions. Eleven volunteers were involved in the process. All of them were native speakers of Tamil with diversity in gender, educational level and medium of instruction in their school education. Table \ref{tab:annotators} shows information about the annotators. The volunteers were instructed to fill up the Google form, and 100 sentences were sent to them. If an annotator offers to volunteer more, the next Google form is sent to them with another set of 100 sentences and in this way each volunteer chooses to annotate as many sentences from the corpus as they want. We send the forms to an equal number of male and female annotators. However, from Table \ref{tab:annotators}, we can see that only two female annotators volunteered to contribute.

\subsection{Corpus Statistics}
 
\begin{table*}[!htb]  
\begin{center} 
\renewcommand{\tabcolsep}{1.5mm}
\captionsetup{font=normalsize}
\normalsize
\begin{tabular}{|l|r|}
\hline
Language pair & Tamil-English \\
\hline
Number of Tokens & 169,833  \\
Vocabulary Size & 30,898\\
Number of Posts &  15,744\\
Number of Sentences & 17,926\\
Average number of Tokens per post & 10 \\
Average number of sentences per post & 1\\
\hline
\end{tabular} 
\caption{Corpus statistic of and Tamil-English} 
\label{tab:corp_stat} 
\end{center} 
\end{table*}
\begin{table}[!htb]  
\begin{center} 
\renewcommand{\tabcolsep}{1.5mm}
\captionsetup{font=normalsize}
\normalsize
\begin{tabular}{|l|r|}
\hline
Class &  Tamil-English \\
\hline
Positive & 10,559 \\
Negative & 2,037\\
Mixed feelings & 1,801 \\
Neutral & 850\\
Other language &  497 \\
\hline
Total & 15,744 \\
\hline
\end{tabular} 
\caption{Data Distribution} 
\label{tab:data_distribution} 
\end{center} 
\end{table}

Corpus statistics is given in the Table \ref{tab:corp_stat}. The distribution of released data is shown in Table \ref{tab:data_distribution}. The entire dataset of 15,744 sentences was randomly shuffled and split into three parts as follows: 11,335 sentences were used for training, 1,260 sentences form the validation set and 3,149 sentences were used for testing. The machine learning models were applied to this subset of data rather than k-fold cross validation. The only other code-mixed dataset of reasonable size that we could find was an earlier work by \newcite{remmiya2016amrita} on code-mix entity extraction for Hindi-English and Tamil-English tweets, released as a part of the shared task in FIRE 2016. The dataset consisted of 3,200 Tanglish tweets used for training and 1,376 tweets for testing. 

\subsection{Inter Annotator Agreement}
We used \textbf{Krippendorff's alpha $(\alpha)$} \cite{krippendorff5} to measure inter-annotator agreement because of the nature of our annotation setup. This is a robust statistical measure that accounts for incomplete data and, therefore, does not require every annotator to annotate every sentence. It is also a measure that takes into account the degree of disagreement between the predicted classes, which is crucial in our annotation scheme. For instance, if the annotators disagree between \textbf{Positive} and \textbf{Negative} class, this disagreement is more serious than when they disagree between \textbf{Mixed feelings} and \textbf{Neutral}. $\alpha$ can handle such disagreements. $\alpha$ is defined as:
\begin{equation}
    \alpha = 1 - \frac{D_o}{D_e}
\end{equation}
$D_o$ is the observed disagreement between sentiment labels by the annotators and $D_e$ is the disagreement expected when the coding of sentiments can be attributed to chance rather than due to the inherent property of the sentiment itself. 
\begin{equation}
    D_o = \frac{1}{n}\sum_{c}\sum_{k}o_{ck\;metric}\;\delta^2_{ck}
\end{equation}

\begin{equation}
D_e = \frac{1}{n(n-1)} \sum_{c}\sum_{k}n_c \; .\;n_{k\;metric}\,\delta^2_{ck}
\end{equation}
Here $o_{ck}\;n_c\;n_k\;$ and $n$ refer to the frequencies of values in coincidence matrices and $metric$ refers to any metric or level of measurement such as nominal, ordinal, interval, ratio and others. Krippendorff's alpha applies to all these metrics. We used nominal and interval metric to calculate annotator agreement. The range of $\alpha$ is between 0 and 1, $1 \ge \alpha \ge 0$. When $\alpha$ is 1 there is perfect agreement between annotators and when 0 the agreement is entirely due to chance. Our annotation produced an agreement of 0.6585 using nominal metric and 0.6799 using interval metric. 
\section{Difficult Examples}\label{section:diff_examples}
In this section we talk about some examples that were difficult to annotate. 
\begin{enumerate}
    \item \textbf{\color{blue}Enakku iru mugan {\color{red}trailer} gnabagam than varuthu} \-- \textit{All it reminds me of is the trailer of the movie Irumugan}. Not sure whether the speaker enjoyed Irumugan trailer or disliked it or simply observed the similarities between the two trailers.
    \item \textbf{\color{blue}Rajini ah vida akshay {\color{red} mass} ah irukane} \-- \textit{Akshay looks more amazing than Rajini}. Difficult to decide if it is a disappointment that the villain looks better than the hero or a positive appreciation for the villain actor. 
    \item \textbf{\color{blue} Ada dei nama sambatha da dei
} \-- \textit{I wonder, Is this our sampath? Hey!.} Conflict between neutral and positive. 
    \item \textbf{\color{blue}Lokesh kanagaraj {\color{red}movie} naalae.... {\color{red}English Rap....Song} vandurum} \-- \textit{If it is a movie of Lokesh kanagaraj, it always has an English rap song}. Ambiguous sentiment.
\end{enumerate}
According to the instructions, questions about music director, movie release date and remarks about when the speaker is watching the video should be treated as neutral. However the above examples show that some comments about the actors and movies can be ambiguously interpreted as neutral or positive or negative. We found annotator disagreements in such sentences.
\section{Benchmark Systems} \label{benchmark_results}
\begin{table*}[!htb] 
\begin{center} 
 \renewcommand{\tabcolsep}{1.5mm}
 \captionsetup{font=small}
 \small
\begin{tabular}{|l|r|r|r|r|r|r|r|r|}
\hline
Classifier & Positive & Negative & Neutral & Mixed & Other language & Micro Avg & Macro Avg & Weighted Avg \\
\hline
KNN & 0.70 & 0.23 & 0.35 & 0.16 & 0.06 & 0.45 & 0.30 & 0.53 \\
Decision Tree & 0.71 & 0.30 & 0.24 & 0.17 & 0.60 & 0.61 & 0.40 & 0.56 \\
Random Forest & 0.69 & 0.51 & 0.80 & 0.41 & 0.68 & 0.68 & 0.62 & 0.63 \\
Logistic Regression & 0.68 & 0.56 & 0.61 & 0.36 & 0.76 & 0.68 & 0.59 & 0.62 \\
Naive Bayes & 0.66 & 0.62 & 0.00 & 0.40 & 0.69 & 0.66 & 0.48 & 0.59 \\
SVM & 0.66 & 0.00 & 0.00 & 0.00 & 0.00 & 0.66 & 0.13 & 0.43 \\
1DConv-LSTM & 0.71 & 0.30 & 0.00 & 0.14 & 0.67 & 0.63 & 0.36 & 0.54 \\
DME & 0.68 & 0.34 & 0.31 & 0.29 & 0.71 & 0.67 & 0.46 & 0.57 \\
CDME & 0.67 & 0.56 & 0.56 & 0.20 & 0.68 & 0.67 & 0.53 & 0.59 \\
BERT Multilingual & 0.67 & 0.00 & 0.00 & 0.00 & 0.64 & 0.67 & 0.26 & 0.46 \\ 
\hline
\end{tabular} 

\caption{Precision } 
\label{tab:precision} 
\end{center} 
\end{table*}
\begin{table*}[!htb] 
\begin{center} 
 \renewcommand{\tabcolsep}{1.5mm}
 \captionsetup{font=small}
 \small
\begin{tabular}{|l|r|r|r|r|r|r|r|r|}
\hline
Classifier & Positive & Negative & Neutral & Mixed & Other language & Micro Avg & Macro Avg & Weighted Avg \\
\hline
KNN & 0.63 & 0.04 & 0.10 & 0.02 & 0.61 & 0.45 & 0.28 & 0.45 \\
Decision Tree & 0.83 & 0.21 & 0.13 & 0.12 & 0.54 & 0.61 & 0.36 & 0.61 \\
Random Forest & 0.98 & 0.18 & 0.09 & 0.04 & 0.55 & 0.68 & 0.32 & 0.68 \\
Logistic Regression & 0.98 & 0.13 & 0.06 & 0.01 & 0.32 & 0.68 & 0.30 & 0.68 \\
Naive Bayes & 1.00 & 0.01 & 0.00 & 0.01 & 0.18 & 0.66 & 0.24 & 0.67 \\
SVM & 1.00 & 0.00 & 0.00 & 0.00 & 0.00 & 0.66 & 0.20 & 0.66 \\
1DConv-LSTM & 0.91 & 0.11 & 0.00 & 0.10 & 0.28 & 0.63 & 0.28 & 0.63 \\
DME & 0.99 & 0.03 & 0.02 & 0.01 & 0.49 & 0.67 & 0.31 & 0.57 \\
CDME & 0.99 & 0.01 & 0.03 & 0.00 & 0.52 & 0.67 & 0.31 & 0.67 \\
BERT Multilingual & 0.99 & 0.00 & 0.00 & 0.00 & 0.58 & 0.67 & 0.31 & 0.46 \\ 
\hline
\end{tabular} 

\caption{Recall} 
\label{tab:recall} 
\end{center} 
\end{table*}

\begin{table*}[!htb] 
\begin{center} 
 \renewcommand{\tabcolsep}{1.5mm}
 \captionsetup{font=small}
 \small
\begin{tabular}{|l|r|r|r|r|r|r|r|r|}
\hline
Classifier & Positive & Negative & Neutral & Mixed & Other language & Micro Avg & Macro Avg & Weighted Avg \\
\hline
KNN & 0.66 & 0.06 & 0.15 & 0.04 & 0.10 & 0.45 & 0.29 & 0.50 \\
Decision Tree & 0.77 & 0.24 & 0.17 & 0.14 & 0.54 & 0.61 & 0.38 & 0.58 \\
Random Forest & 0.81 & 0.18 & 0.09 & 0.04 & 0.55 & 0.68 & 0.42 & 0.65 \\
Logistic Regression & 0.81 & 0.21 & 0.12 & 0.03 & 0.45 & 0.68 & 0.40 & 0.64 \\
Naive Bayes & 0.80 & 0.02 & 0.00 & 0.01 & 0.29 & 0.66 & 0.32 & 0.63 \\
SVM & 0.79 & 0.00 & 0.00 & 0.00 & 0.00 & 0.66 & 0.16 & 0.52 \\
1DConv-LSTM & 0.80 & 0.16 & 0.00 & 0.12 & 0.39 & 0.63 & 0.31 & 0.58 \\
DME & 0.80 & 0.05 & 0.04 & 0.01 & 0.58 & 0.67 & 0.37 & 0.57 \\
CDME & 0.80 & 0.02 & 0.05 & 0.01 & 0.59 & 0.67 & 0.39 & 0.63 \\
BERT Multilingual & 0.80 & 0.00 & 0.00 & 0.00 & 0.61 & 0.67 & 0.28 & 0.46 \\ 
\hline
\end{tabular} 

\caption{F-score} 
\label{tab:f-score} 
\end{center} 
\end{table*}

In order to provide a simple baseline, we applied various machine learning algorithms for determining the sentiments of YouTube posts in code-mixed Tamil-English language. 
\subsection{Experimental Settings}\label{subsection:exp_settings}
\subsubsection{Logistic Regression (LR):} We evaluate the Logistic Regression model with L2 regularization. The input features are the Term Frequency Inverse Document Frequency (TF-IDF) values of up to 3 grams.
\subsubsection{Support Vector Machine (SVM):} We evaluate the SVM model with L2 regularization. The features are the same as in LR. The purpose of SVM classification algorithm is to define optimal hyperplane in N dimensional space to separate the data points from each other.
\subsubsection{K-Nearest Neighbour (K-NN):} We use KNN for classification with 3,4,5,and 9 neighbours by applying uniform weights.
\subsubsection{Decision Tree (DT):} Decision trees have been previously used in NLP tasks for classification. In decision tree, the prediction is done by splitting the root training set into subsets as nodes, and each node contains output of the decision, label or condition. After sequentially choosing alternative decisions, each node recursively is split again and finally the classifier defines some rules to predict the result.  We used it to classify the sentiments for baseline. Maximum depth  was 800 and minimum sample splits were 5 for DT. The criterion were Gini and entropy.
\subsubsection{Random Forest (RF):} In random forest, the classifier randomly generates trees without defining rules. We evaluate the RF model with same features as in DT. 
\subsubsection{Multinominal Naive Bayes (MNB):} Naive-Bayes classifier is a probabilistic model, which is derived from Bayes Theorem that finds the probability of hypothesis activity to the given evidence activity. We evaluate the MNB model with our data using $\alpha$=1 with TF-IDF vectors. 
\subsubsection{1DConv-LSTM:} The model we evaluated consists of Embedding layer, Dropout, 1DConv with activation ReLU, Max-pooling and LSTM. The embeddings are randomly initialized.
\subsubsection{BERT-Multilingual:} \newcite{devlin-etal-2019-bert} introduced a language representation model which is Bidirectional Encoder Representation from Transforms. It is designed to pre-train from unlabelled text and can be fine-tuned by adding last layer. BERT has been used for many text classification tasks \cite{tayyar-madabushi-etal-2019-cost,ma-etal-2019-domain,cohan-etal-2019-pretrained}. We explore classification of a code-mixed data into their corresponding sentiment categories.
\subsubsection{DME and CDME:} We also implemented the Dynamic Meta Embedding \cite{kiela-etal-2018-dynamic} to evaluate our model. As a first step, we used Word2Vec and FastText to train from our dataset  since dynamic meta-embedding is an effective method for the supervised learning of embedding ensembles.
\subsection{Experiment Results and Discussion}
The experimental results of the sentiment classification task using different methods are shown in terms of precision in Table \ref{tab:precision}, recall in Table \ref{tab:recall}, and F-score in Table \ref{tab:f-score}. We used \textit{sklearn} \footnote{https://scikit-learn.org/} for evaluation. The micro-average is calculated by aggregating the contributions of all classes to compute the average metric. In a multi-class classification setup, micro-average is preferable if there are class imbalances. For instance in our data, we have many more examples of positive classes than other classes. A macro-average will compute the metrics (precision, recall, F-score) independently for each class and then take the average. Thus this metric treats all classes equally and it does not take imbalance into account. A weighted average takes the metrics from each class just like macro but the contribution of each class to the average is weighted by the number of examples available for it. For our test, positive is 2,075, negative is 424, neutral is 173, mixed feelings are 377, and non-Tamil is 100. \\

As shown in the tables, all the classification algorithms perform poorly on the code-mixed dataset. Logistic regression, random forest classifiers and decision trees were the ones that fared comparatively better across all sentiment classes. Surprisingly, the classification result by the SVM model has much worse diversity than the other methods. Applying deep learning methods also does not lead to higher scores on the three automatic metrics. We think this stems from the characteristics of the dataset. The classification scores for different sentiment classes appear to be in line with the distribution of sentiments in the dataset.\\

The dataset is not a balanced distribution. Table \ref{tab:data_distribution} shows that out of total 15,744 sentences ~67\% belong to \textit{Positive} class while the other sentiment classes share ~13\%,~5\% and ~3\% respectively. The precision, recall and F-measure scores are higher for the \textit{Positive} class while the scores for \textit{Neutral} and \textit{Mixed feeling} classes were disastrous. Apart from their low distribution in the dataset, these two classes are difficult to annotate for even human annotators as discussed in Section \ref{section:diff_examples} In comparison, the \textit{Negative} and \textit{Other language} classes were better. We suspect this is due to more explicit clues for negative and non-Tamil words and due to relatively higher distribution of negative comments in the data.\\

Since we collected the post from movie trailers, we got more positive sentiment than others as the people who watch trailers are more likely to be interested in movies and this skews the overall distribution. However, as the code-mixing phenomenon is not incorporated in the earlier models, this resource could be taken as a starting point for further research. There is significant room for improvement in code-mixed research with our dataset. In our experiments, we only utilized the machine learning methods, but more information such as linguistic information or hierarchical meta-embedding can be utilized. This dataset can be used to create a multilingual embedding for code-mixed data \cite{pratapa-etal-2018-word}.
\section{Conclusion}
We presented, to the best of our knowledge, the most substantial corpus for under-resourced code-mixed Tanglish with annotations for sentiment polarity. We achieved a high inter-annotator agreement in terms of Krippendorff $\alpha$ from voluntary annotators on contributions collected using Google form. We created baselines with gold standard annotated data and presented our results for each class in Precision, Recall, and F-Score. We expect this resource will enable the researchers to address new and exciting problems in code-mixed research. 
\section{Acknowledgments}
This publication has emanated from research supported in part by a research grant from Science Foundation Ireland (SFI) under Grant Number SFI/12/RC/2289 (Insight), SFI/12/RC/2289$\_$P2 (Insight$\_$2), co-funded by the European Regional Development Fund as well as by the EU H2020 programme under grant  agreements 731015 (ELEXIS-European Lexical Infrastructure), 825182 (Prêt-à-LLOD), and Irish Research Council grant IRCLA/2017/129 (CARDAMOM-Comparative Deep Models of Language for Minority and Historical Languages).
\section{Bibliographical References}
\bibliographystyle{lrec.bst}
\bibliography{lrec2020W-xample-kc}

\end{document}